\documentclass[10pt,twocolumn,letterpaper]{article}
\usepackage{authblk}

\usepackage{algorithmicx}
\usepackage{algorithm}
\usepackage{algpseudocode}
\usepackage{multirow}
\usepackage{pifont}
\usepackage[accsupp]{axessibility}
\usepackage[applications]{wacv}      

\definecolor{wacvblue}{rgb}{0.21,0.49,0.74}
\usepackage{chapterbib}
\usepackage[pagebackref,breaklinks,colorlinks,allcolors=wacvblue]{hyperref}

\def\wacvPaperID{36} 
\def\confName{WACV}
\def\confYear{2027}

\title{UniPrompt-CL: Efficient Single-Inference Continual Learning for Medical Image Classification}

\author{
Gyutae Oh \qquad Jitae Shin\thanks{Corresponding author.}\\
Department of Electrical and Computer Engineering\\
Sungkyunkwan University\\
Suwon 16419, Republic of Korea\\
{\tt\small alswo740012@g.skku.edu \qquad jtshin@skku.edu}
}
\begin{document}
\begin{cbunit}
\maketitle
{\renewcommand{\thefootnote}{}\footnotetext{Code: \href{https://github.com/GYUGYUT/UniPrompt-CL}{\raisebox{-0.2em}{\includegraphics[height=1em]{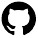}}\,{\urlstyle{same}\url{https://github.com/GYUGYUT/UniPrompt-CL}}}}\addtocounter{footnote}{-1}}

\begin{abstract}
Medical AI systems must remain reliable as data distributions shift across hospitals, imaging devices, and patient populations, yet privacy constraints often prevent storing or revisiting past patient data. We present UniPrompt-CL, a rehearsal-free domain-incremental learning framework for medical image classification using a single backbone and a single inference. Rather than relying on multiple backbones, repeated ViT inferences, or independent layer-wise prompt pools, UniPrompt-CL shares a unified prompt pool across transformer layers and expands it with only a few prompts per new domain. This reduces cross-layer redundancy while enabling efficient adaptation. On diabetic retinopathy, UniPrompt-CL improves average accuracy by 2.0 percentage points and substantially reduces forgetting relative to the strongest backbone-matched baseline. On skin cancer, it achieves competitive accuracy at the lowest training cost among DINOv2-B configurations. A controlled comparison varying only pool organization shows that the benefit arises from retention rather than learning each new domain: it is largest on the medical benchmarks, smaller on OfficeHome, and reverses on a heterogeneous six-stage DomainNet stream. These results suggest that cross-layer prompt sharing with lightweight expansion is practical for medical models requiring repeated retraining under limited on-site compute, while identifying the domain shifts under which it helps.

\end{abstract}

\begin{figure}[t]
\centering
\includegraphics[width=\columnwidth]{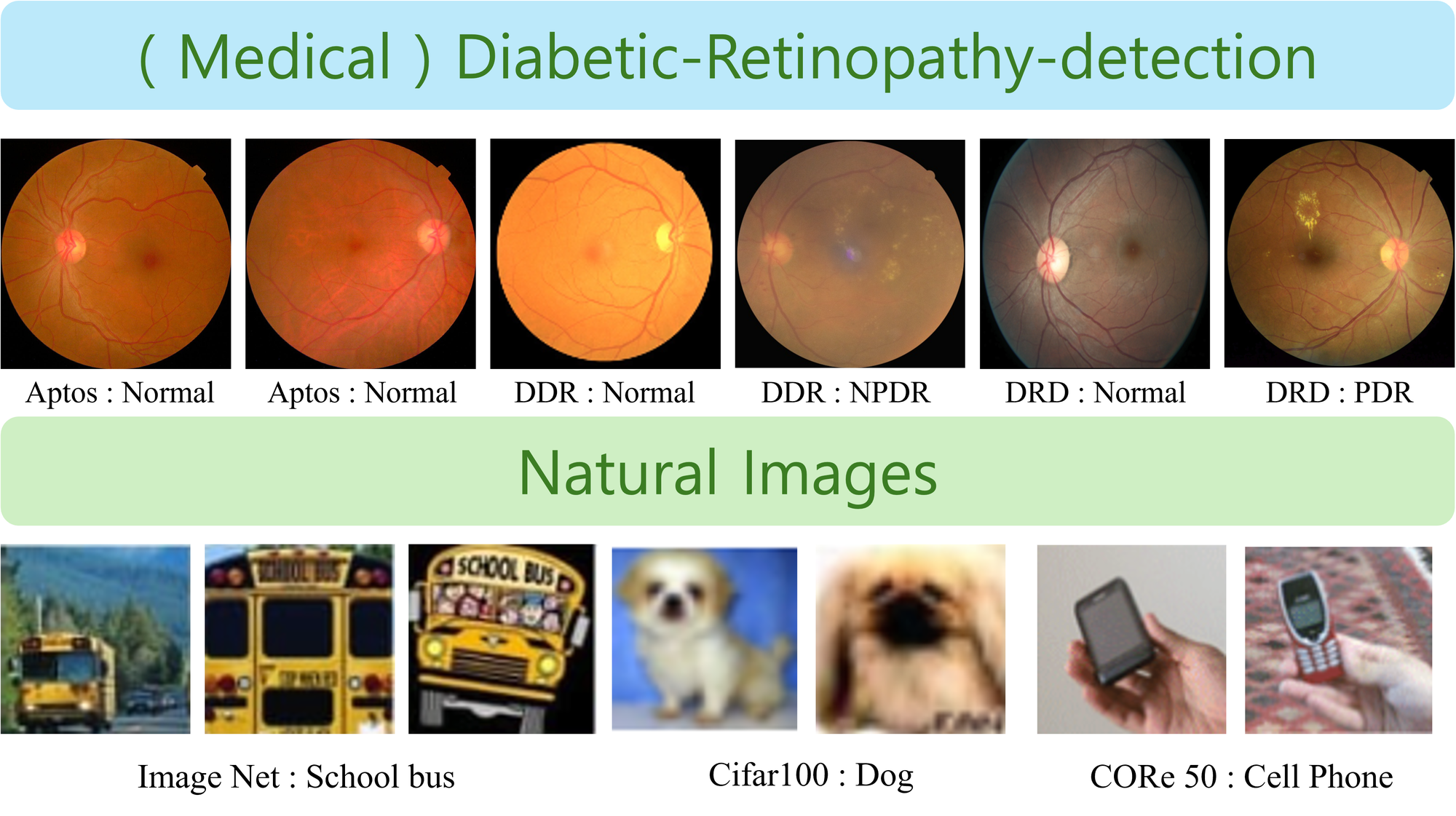}
\caption{Representative examples of natural images used in prior PCL studies and
medical images used in this study, highlighting the standardized acquisition conditions
and fine-grained visual differences in medical datasets.}
\label{figure2}
\end{figure}

\section{Introduction}
\label{sec:intro}

Medical artificial intelligence systems are increasingly expected to operate beyond a single static training distribution. In clinical practice, image data arrive sequentially from different hospitals, imaging devices, acquisition protocols, and patient populations, while privacy and institutional constraints often prevent storing or revisiting previously observed patient data. These conditions make rehearsal-free domain-incremental learning (DIL) an important setting for medical AI that must be updated repeatedly after deployment. Yet most continual learning (CL) methods have been developed and validated on natural-image benchmarks, and thus do not fully reflect key characteristics of medical data such as domain bias, inter-device heterogeneity, subtle inter-stage boundaries, and severe class imbalance.

The variation that prompt-based continual learning (PCL) must capture differs in character
between the two domains. We do not claim that medical images are universally harder;
rather, the relevant axis of variation is different. As shown in Fig.~\ref{figure2}, the
medical datasets we evaluate (APTOS~\cite{aptos2019-blindness-detection},
DDR~\cite{LI2019}, DRD~\cite{diabetic-retinopathy-detection}) are collected under
relatively standardized acquisition protocols, yielding more consistent anatomical
structures and capture angles than the natural-image benchmarks used in prior PCL studies
(e.g., ImageNet~\cite{imagenet-object-localization-challenge},
CIFAR100~\cite{Krizhevsky09learningmultiple},
CORe50~\cite{lomonaco2017core50})~\cite{chen2024promptfusion,wang2022learning,smith2023coda,kim2024one}.
The clinically meaningful differences instead appear as subtle, localized changes in
color, texture, and lesion appearance---for example, between DDR non-proliferative (NPDR)
and DRD proliferative (PDR) diabetic retinopathy (DR). Under this regime, limited prompt
capacity is better spent compactly modeling fine-grained domain-specific variation than
broadly covering large appearance changes; the skin-cancer examples in
Sup.~Fig.~7 are consistent with this.

Natural images, by contrast, exhibit large intra-class variability in viewpoint, shape,
background, and appearance, so PCL methods designed for them often learn widely
distributed, layer-wise prompt regions. Under the more
standardized medical shifts we consider, where fundamental anatomical structure stays
consistently important across layers, such layer-wise independent pools repeatedly encode
similar low-level information, causing cross-layer redundancy and inefficient use of
limited prompt capacity. Many PCL methods additionally rely on multiple backbones or
repeated Vision Transformer (ViT) inferences (e.g., a second forward pass or extra query
generation), which raises the cost of every adaptation
round~\cite{wang2022learning,smith2023coda,menabue2024semantic}. This matters
because data-governance constraints may require the update for a new site,
device, or acquisition protocol to be carried out on local compute, so the
binding constraint is the training capacity of an on-site workstation rather
than inference latency at the bedside. We treat these
visualizations as qualitative motivation and validate the resulting design in
Sec.~\ref{sec:experiments}.

Motivated by these observations, we propose UniPrompt-CL, a single-pass prompt-based
continual learning framework for rehearsal-free domain-incremental medical image
classification. UniPrompt-CL avoids replaying past patient data, operates with a single
backbone, and requires only a single ViT inference. To reduce cross-layer redundancy while
preserving adaptability, it consolidates layer-wise prompt pools into a unified pool and
expands only a few prompts at each new stage, retaining previously learned knowledge at
low computational overhead.

We evaluate UniPrompt-CL on two medical domain-incremental settings: a
within-domain sequence on DR
(APTOS~$\rightarrow$~DDR~$\rightarrow$~DRD) and a separate sequence on skin
cancer (ISIC~$\rightarrow$~HAM~$\rightarrow$~DERM7) from a different medical
domain. Across both, UniPrompt-CL achieves stronger
accuracy--forgetting--cost trade-offs than representative PCL,
rehearsal-based, regularization-based, and architecture-based baselines. To
test how far the design carries, we then vary only the pool organization and
add two natural-image sequences, OfficeHome and a six-stage DomainNet stream,
which locate the shift structures under which cross-layer sharing helps and
those under which it does not.

\paragraph{\textbf{Contributions.}}
\begin{itemize}
\item We formulate a rehearsal-free medical DIL setting that reflects practical deployment constraints, including institutional domain shifts, privacy limitations, and the inability to revisit past patient data.

\item We propose UniPrompt-CL, a single-backbone, single-inference PCL framework that replaces layer-wise independent prompt pools with a unified prompt pool shared across transformer layers.

\item We introduce a lightweight stage-wise prompt expansion strategy with a surrogate selection regularization term, enabling newly added prompts to participate in adaptation while preserving previously learned prompt knowledge.

\item We show, in a controlled comparison that varies only the pool organization, that the benefit of cross-layer sharing depends on the structure of the domain shift: it is largest under the standardized acquisition shifts of our medical benchmarks, shrinks on OfficeHome, and reverses on the more heterogeneous six-stage DomainNet stream, so the advantage is shift-dependent rather than universal.

\item We validate UniPrompt-CL on DR and skin-cancer benchmarks, demonstrating improved accuracy on DR and competitive accuracy on skin cancer, together with reduced forgetting and lower computational cost under backbone-controlled comparisons.
\end{itemize}

\section{Related Work}
\label{sec:related_work}

\paragraph{Continual Learning for Medical AI Deployment.}
CL enables models to acquire new knowledge over time while retaining
previously learned information~\cite{wang2024comprehensive,oh2025multi}. This is especially relevant
to medical AI, where distributions shift across hospitals, devices, protocols, and patient
populations, and where static models degrade under domain shift while past patient data
often cannot be stored or revisited under institutional and privacy
constraints~\cite{qazi2024continual}. We therefore study a rehearsal-free DIL setting in which
new datasets arrive sequentially and earlier-stage data is inaccessible. A related privacy-preserving paradigm is federated learning (FL), which supports multi-institutional training without sharing patient data \cite{sheller2020federated,rieke2020future,shin2026causal}.
FL and rehearsal-free DIL address orthogonal axes,
however: FL assumes that participating sites coexist and remain accessible throughout a coordinated training protocol, whereas our setting is temporal---domains arrive as a stream and past data cannot be revisited once a stage has passed. FL thus does not by
itself address forgetting along a sequential stream, and is complementary to, rather than a substitute for, the continual setting we study (e.g., FL may operate within each stage).
We accordingly evaluate against continual learning baselines.

\paragraph{Rehearsal-Free and Parameter-Efficient Continual Learning.}
Existing CL methods are commonly grouped into regularization-, architecture-, and
rehearsal-based approaches. Regularization-based methods such as EWC~\cite{kirkpatrick2017overcoming} constrain updates by estimating the importance of previously learned weights, but these approximations can become unreliable
as data complexity and model dimensionality grow~\cite{wang2024toward,pham2022continual,zenke2017continual}.
Architecture-based methods allocate task-specific parameters, subnetworks, or
masks~\cite{ebrahimi2020adversarial,wortsman2020supermasks}, yet their cost can grow with
the number of stages, limiting long-term deployment. Rehearsal-based methods such as Dark Experience Replay~\cite{buzzega2020dark} replay stored examples~\cite{wu2018memory,shin2017continual,rebuffi2017icarl}, but storing real patient data is
often undesirable or infeasible in privacy-sensitive settings. These limitations motivate
rehearsal-free, parameter-efficient alternatives for medical CL.

\paragraph{Prompt-Based Continual Learning.}
PCL freezes most of a pretrained backbone and adapts
only a small set of learnable prompts, preserving pretrained representations while adapting
to new domains at low memory cost~\cite{chen2024promptfusion,wang2022learning,smith2023coda,kim2024one}.
As noted above, however, most PCL methods are validated on natural images and rely on
additional query functions, multiple backbones, or repeated ViT inferences, limiting their
suitability when each adaptation round must run on local hospital compute~\cite{wang2022learning,smith2023coda,menabue2024semantic}.
Recent work targets training-time efficiency: CPS-Prompt~\cite{lim2026critical} combines
critical patch sampling with decoupled prompt/classifier training to cut on-device memory
and compute while remaining competitive. Such methods reduce per-pass cost but retain the
two-stage query--prompt pipeline, and are thus orthogonal and complementary to our goal of
removing repeated inference altogether; we include CPS-Prompt as a contemporary
efficiency-oriented baseline.

\paragraph{Positioning of This Work.}
Our work bridges PCL and practical medical AI deployment.
Unlike rehearsal-based methods, UniPrompt-CL does not store or replay past medical images;
unlike architecture-expanding approaches, it uses a single backbone and a single ViT
inference; and unlike prior PCL methods with layer-wise independent prompt pools, it
consolidates prompts into a unified pool and expands only a small subset at each stage.
This design follows from the observation that the shifts we target are
sufficiently fine-grained that adaptation can be reused across layers rather
than allocated separately at each one. UniPrompt-CL is thus a rehearsal-free PCL framework for
domain-incremental medical image classification with a low retraining cost, where
the operative constraint is the inability to revisit past data rather than the
distribution of training across sites. Sec.~\ref{sec:controlled} then asks how far
this design carries beyond that setting.

\begin{figure*}[t]
\centering
\includegraphics[width=0.8\textwidth, keepaspectratio]{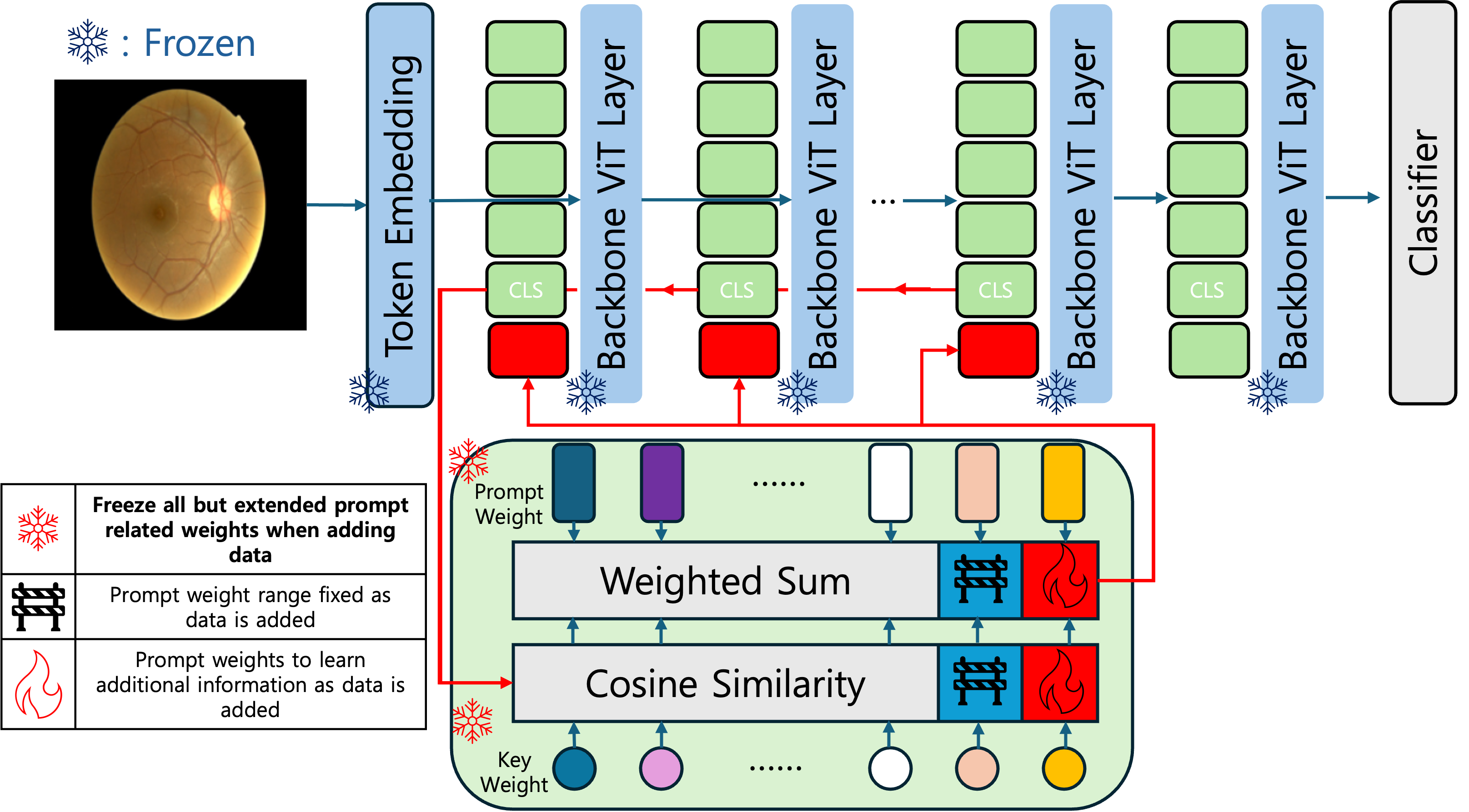}
\caption{Overall architecture. At each layer the [CLS] token serves as a query;
the prompt-pool integration module retrieves a prompt from the unified pool and
fuses it with the layer representation. At each new stage a small number of
prompts is added, while all previously learned prompts remain frozen.}
\label{figure1}
\end{figure*}

\section{Proposed Method}
\label{sec:Proposed Method}

\subsection{Preliminary and Overview}
\label{sec:preliminary}
PCL freezes a pretrained ViT backbone and adapts only a
small set of learnable prompt vectors inserted into intermediate layers. To select which
prompts to use, conventional PCL relies on a query function---often a separate ViT---that
encodes the input into a query matched against prompt keys, incurring an extra forward pass
and substantial overhead. Our method builds upon the One-Stage PCL framework
(OS-Prompt)~\cite{kim2024one}, which removes this auxiliary function. In a ViT, each layer
carries a [CLS] token that aggregates global image information and is conventionally used
for classification at the final layer; OS-Prompt reuses this intermediate [CLS] token
directly as the prompt query. Following OS-Prompt, we use the layer-wise [CLS] token
as the prompt query:
\begin{equation}
q_l = x_{l}[\text{CLS}], \quad q_l \in \mathbb{R}^{1 \times D},
\label{eq:query}
\end{equation}
where $x_l \in \mathbb{R}^{N \times D}$ denotes the token embedding at layer $l$, $N$ is
the number of tokens, and $D$ is the embedding dimension. A prompt is then retrieved from
the pool via a similarity-weighted sum over prompt keys, as detailed in
Sup.~A. A single intermediate [CLS] token has limited representational capacity; its extension, OS-Prompt++, recovers expressiveness but requires two ViT inferences during training, and thus is not a true single-inference solution.

While OS-Prompt primarily addresses the efficiency of query generation, our work addresses
a different problem: how prompts should be organized, expanded, and regularized for medical
DIL. We retain the single-backbone query mechanism of OS-Prompt,
but replace its layer-wise independent prompt pools with a unified prompt pool, and
introduce a lightweight prompt-expansion strategy with selection regularization. A more
detailed comparison between OS-Prompt, OS-Prompt++, and our method is provided in
Sup.~A. Our goal is to achieve superior DIL
with a single backbone and a single inference, through three components: (1) a unified
prompt pool, (2) a few-prompt expansion strategy, and (3) a surrogate prompt-selection
regularization term. Because medical datasets often exhibit severe long-tailed
distributions that aggravate catastrophic forgetting, we adopt the
\textbf{DINOv2-base (86.6M parameters)}~\cite{oquab2023dinov2} backbone for its balance of
representational capacity and efficiency; dataset statistics and the backbone comparison
are provided in Sup.~C. The overall architecture is illustrated in
Fig.~\ref{figure1}, and the full algorithm in Sup.~D
(Algorithm~1).

\subsection{Integration of Prompt-Pool}
\label{sec:integration}
\textbf{Motivation.} As discussed in Section~\ref{sec:intro}, the
domain-incremental shifts we target call for compactly modeling subtle,
fine-grained variations rather than broadly dispersed prompt representations;
Sec.~\ref{sec:controlled} examines how far that assumption carries. Layer-wise independent
prompt pools instead encode similar low-level information redundantly across layers, spending limited prompt capacity inefficiently. We therefore share a single unified prompt pool across all layers, which removes this cross-layer redundancy and lets the available prompt capacity be reused consistently throughout the backbone. This is not only a qualitative observation: measuring OS-Prompt's trained pools,
prompts and keys carrying the same index in different layers have mean absolute
cosine similarity $0.990$ and $0.988$, whereas different prompts within a single
layer are near-orthogonal ($0.029$ and $0.030$). Each layer therefore keeps its
prompts internally diverse while duplicating the same directions across layers.
The full analysis is given in Sup.~B.

\noindent\textbf{Method.} Using the layer [CLS] token as query $q_l$ (Eq.~\ref{eq:query}), we
define unified prompt keys $k^m \in \mathbb{R}^{1 \times D}$ and values
$p^m \in \mathbb{R}^{1 \times D}$ that are shared across all layers. The prompt injected
at layer $l$ is a weighted combination of the prompt values,
\begin{equation}
\phi_l =
\sum_{m=1}^{L_p}
\gamma(q_l, k^m)\, p^m,
\quad
\phi_l \in \mathbb{R}^{1 \times D},
\label{eq:unified}
\end{equation}
where $\gamma(q_l, k^m)$ is a similarity-based weight between the query and the
$m$-th key, and $L_p$ is the number of prompts in the unified pool. The prompt is then
injected back into the backbone: $f_l(\cdot)$ denotes the frozen $l$-th transformer block,
which takes the layer input $x_l$ together with the prompt $\phi_l$ and produces the next
layer representation,
\begin{equation}
x_{l+1} = f_l\bigl(x_l, \phi_l\bigr).
\label{eq:fuse}
\end{equation}

\subsection{Few Prompt Expansion}
\label{sec:expansion}
In this subsection, we introduce the Few Prompt Expansion strategy. Building on the single-inference OS-Prompt mechanism, this strategy maximizes
the use of existing prompt information while providing additional capacity
to acquire new domain knowledge, without introducing extra ViT inference. When
transitioning to the next CL stage, we freeze the prompt weights corresponding to the
previous stages and expand the prompt set by adding a fixed fraction $\rho$ of the
original $L_p$ prompts, with $\rho=20\%$ in the main experiments. This ratio was
empirically determined in Fig.~\ref{fig:prompt_expansion_ratio}. The number of
newly added prompts is denoted as $\psi=\lfloor \rho L_p \rfloor$, resulting in
an expanded prompt pool of size
$\hat{L}_p = L_p + \psi$. We write $\mathcal{I}_{\psi} = \{L_p+1,\dots,\hat{L}_p\}$ for the
index set of the newly added prompts, and $\mathcal{I}$ for the set of prompted layers.

Since the newly added prompts may initially have less influence than well-trained
previous prompts, they can be under-utilized during the early phase of domain expansion.
To address this issue, we introduce a surrogate selection loss that encourages the newly
added prompts to receive higher selection scores, i.e., to be retrieved more often
from the unified pool. This loss is not designed to enforce orthogonality between new and previous prompts; instead, it counteracts the early-stage under-selection of newly added prompts and allows them to be trained on the new domain.

For the $b$-th input sample at layer $l$, let $q_{b,l}$ denote the layer-wise [CLS]
query and $B$ the batch size. Given the expanded prompt-key set
$\{k^m\}_{m=1}^{\hat{L}_p}$, we compute the normalized selection score of prompt $m$ as
\begin{equation}
z_{b,l,m} =
\mathrm{Softmax}_{m}
\left(
\frac{q_{b,l}(k^m)^{\top}}
{\|q_{b,l}\|_2 \|k^m\|_2}
\right),
\quad
m=1,\dots,\hat{L}_p,
\label{eq:z}
\end{equation}
where the softmax is taken over the $\hat{L}_p$ prompts, so $z_{b,l,m} \in (0,1)$ is the
relative probability of selecting prompt $m$ for query $q_{b,l}$. We then define a
per-prompt selection penalty for each newly added prompt,
\begin{equation}
S_{b,l,m} = 1 - z_{b,l,m},
\quad m \in \mathcal{I}_{\psi},
\label{eq:selection_penalty}
\end{equation}
which is small when a new prompt receives a high selection score, and aggregate it into
the prompt expansion loss
\begin{equation}
\mathcal{L}_s =
\frac{1}{B\,|\mathcal{I}|\,|\mathcal{I}_{\psi}|}
\sum_{b=1}^{B}
\sum_{l \in \mathcal{I}}
\sum_{m \in \mathcal{I}_{\psi}}
S_{b,l,m},
\label{eq:Ls}
\end{equation}
where the normalization is by the batch size $B$, the number of prompted layers
$|\mathcal{I}|$, and the number of newly added prompts $|\mathcal{I}_{\psi}|$. Because the
softmax in Eq.~\eqref{eq:z} is normalized over all $\hat{L}_p$ prompts while
$\mathcal{L}_s$ penalizes only the new subset $\mathcal{I}_{\psi}$, the loss does not
collapse into a constant: lowering it moves selection probability mass toward the
newly added prompts. Minimizing $\mathcal{L}_s$ is thus intended to keep the newly
introduced prompts from being ignored during the early stage of domain expansion. We note that a higher selection score
does not by itself guarantee that a prompt contributes to the final prediction; rather,
$\mathcal{L}_s$ ensures the new prompts are actually exercised on the new domain, and
their ultimate usefulness is reflected in the downstream classification objective below
and in the ablation in Sec.~\ref{sec:ablation}.

Finally, the overall loss function is defined as
\begin{equation}
\mathcal{L}_{\mathrm{total}} =
\mathcal{L}_{\mathrm{CE}} +
\lambda \mathcal{L}_s,
\label{eq:total}
\end{equation}
where $\lambda$ is set to 0.001 in this study to control the strength of the
regularization term. The effect of $\lambda$ is further analyzed in the ablation study.

\begin{table*}[t]
\centering
\footnotesize
\setlength{\tabcolsep}{10pt}
\renewcommand{\arraystretch}{0.90}
\caption{Continual learning performance and training cost. $*$ marks a DINOv2-B backbone, added for the applicable prompt-based baselines so that backbone-level and method-level gains can be separated; the others use their original configuration. GFLOPs is the per-step training cost (forward and backward) and depends on the method and backbone, not the benchmark. BWT is omitted: it coincides with $-\mathrm{AvgF}$ throughout these sequences (Sup.~F). $\dagger$ denotes the proposed model; \textbf{bold} is the best value per benchmark.}
\label{Table3}
\begin{tabular}{l c c c c c}
\toprule
\multirow{2}{*}{\textbf{Method}} &
\multirow{2}{*}{\textbf{GFLOPs}$\downarrow$} &
\multicolumn{2}{c}{\textbf{Diabetic Retinopathy}} &
\multicolumn{2}{c}{\textbf{Skin Cancer}} \\
\cmidrule(lr){3-4} \cmidrule(lr){5-6}
& &
\textbf{AvgACC}$\uparrow$ & \textbf{AvgF}$\downarrow$ &
\textbf{AvgACC}$\uparrow$ & \textbf{AvgF}$\downarrow$ \\
\midrule
OS-Prompt~\cite{kim2024one}
& \textbf{34.26}
& 0.666$\pm$0.062 & 0.132$\pm$0.003
& 0.682$\pm$0.041 & 0.135$\pm$0.075 \\
MoE-Adapters~\cite{yu2024boosting}
& 105.64
& 0.716$\pm$0.141 & 0.080$\pm$0.092
& 0.597$\pm$0.165 & 0.040$\pm$0.016 \\
Coda-Prompt~\cite{smith2023coda}
& 134.33
& 0.688$\pm$0.079 & 0.140$\pm$0.044
& 0.713$\pm$0.040 & 0.041$\pm$0.013 \\
DualPrompt~\cite{wang2022dualprompt}
& 105.05
& 0.467$\pm$0.159 & 0.291$\pm$0.065
& 0.637$\pm$0.065 & \textbf{0.012}$\pm$0.039 \\
DER++~\cite{buzzega2020dark}
& 168.02
& 0.607$\pm$0.121 & 0.288$\pm$0.160
& 0.722$\pm$0.057 & 0.099$\pm$0.023 \\
Online EWC~\cite{schwarz2018progress}
& 100.62
& 0.715$\pm$0.151 & 0.174$\pm$0.002
& 0.708$\pm$0.054 & 0.157$\pm$0.002 \\
OS-Prompt++~\cite{kim2024one}
& 51.12
& 0.769$\pm$0.027 & 0.113$\pm$0.006
& 0.725$\pm$0.032 & 0.063$\pm$0.013 \\
CPS-Prompt~\cite{lim2026critical}
& 70.29
& 0.769$\pm$0.004 & 0.153$\pm$0.037
& 0.680$\pm$0.024 & 0.086$\pm$0.032 \\
\midrule
CPS-Prompt$^*$
& 92.69
& 0.692$\pm$0.001 & 0.116$\pm$0.017
& 0.720$\pm$0.015 & 0.087$\pm$0.035 \\
OS-Prompt$^*$
& 66.42
& 0.801$\pm$0.005 & 0.165$\pm$0.006
& 0.726$\pm$0.011 & 0.094$\pm$0.021 \\
OS-Prompt++$^*$
& 86.68
& 0.812$\pm$0.023 & 0.125$\pm$0.052
& 0.724$\pm$0.011 & 0.103$\pm$0.020 \\
Coda-Prompt$^*$
& 175.12
& 0.824$\pm$0.003 & 0.154$\pm$0.024
& 0.716$\pm$0.024 & 0.109$\pm$0.036 \\
DualPrompt$^*$
& 137.93
& 0.733$\pm$0.009 & 0.104$\pm$0.018
& 0.728$\pm$0.002 & 0.079$\pm$0.007 \\
\midrule
\textbf{UniPrompt-CL}$^{\dagger*}$
& 44.17
& \textbf{0.844}$\pm$0.016 & \textbf{0.079}$\pm$0.032
& \textbf{0.732}$\pm$0.022 & 0.049$\pm$0.043 \\
\bottomrule
\end{tabular}
\end{table*}

\section{Experiment Settings}
In this section, we describe our experimental setup in detail to substantiate our claims.
We first introduce the datasets used in our experiments and then summarize the data
augmentation, preprocessing, baseline validation, evaluation metrics, and implementation
environment. Detailed information about the datasets and experimental settings is
provided in Sup.~E.

\noindent\textbf{Continual Learning Problem Setting \& Dataset:}
We simulate a hospital-driven DIL setting with unknown
task identities, following a rehearsal-free protocol in which data from past or future stages are inaccessible~\cite{wang2024comprehensive}.  Our main experiments are conducted on DR datasets, with domains presented in the incremental order APTOS~\cite{aptos2019-blindness-detection}
$\rightarrow$ DDR~\cite{LI2019} $\rightarrow$ DRD~\cite{diabetic-retinopathy-detection}. To further assess generalization, we additionally evaluate on skin cancer datasets (ISIC~\cite{rotemberg2021patient} $\rightarrow$ HAM~\cite{codella2019skin} $\rightarrow$ DERM7~\cite{kawahara2018seven}). Example images and additional
dataset details are provided in Sup.~E.

\noindent\textbf{Experimental Details:}
During training, we apply random data augmentations, including HorizontalFlip, Rotation, and VerticalFlip. For ViT preprocessing, we use AutoImageProcessor from the transformers library with facebook/dinov2-base (DINOv2-B) weights. We validate baseline methods using rigorously implemented continual learning libraries~\cite{boschini2022class,buzzega2020dark}, and implement any additional required components ourselves. For a rigorous evaluation, we report the standard continual learning metrics
AvgACC$\uparrow$ and AvgF$\downarrow$ together with the raw training cost in
GFLOPs, so that predictive performance, retention, and computation can each be
read separately rather than through a single compound score. AvgACC is the
average performance over the whole training trajectory and AvgF the average
performance lost on earlier domains after the final stage; full definitions are given in Sup.~F.

\section{Experiment Results}
\label{sec:experiments}
In this section, we present the experimental results by grouping each table and figure
according to its role in validating the proposed method.

\begin{figure}[t]
\centering
\includegraphics[width=0.7\columnwidth]{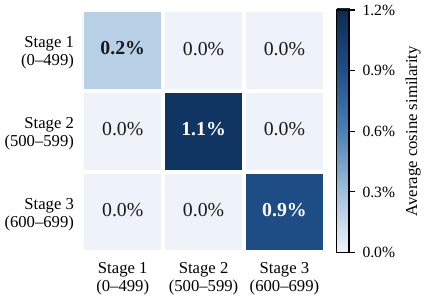}
\caption{Inter-stage similarity of incrementally added prompts. Newly added
prompts exhibit limited overlap with prompts from earlier stages, consistent
with each expansion contributing complementary stage-specific directions rather
than duplicating existing ones. We treat this as qualitative motivation,
validated empirically in Sec.~\ref{sec:experiments}.}
\label{figure3}
\end{figure}

\subsection{Stage-Wise Prompt Expansion and Redundancy}
As motivated in Section~\ref{sec:expansion}, prompts introduced at each stage should
preserve previously acquired knowledge while encoding new domain-specific information.
Fig.~\ref{figure3} visualizes the inter-stage prompt similarity matrix under our
method. The diagonal entries are consistently larger than the off-diagonal
entries, while cross-stage similarities remain near zero. This is consistent
with limited overlap between newly added prompts and those from earlier
stages. We interpret this as evidence of
reduced redundancy across stages; it does not by itself establish that each prompt is
used for the downstream task, which is instead assessed through the classification
performance below.

The cross-layer counterpart of this observation is the redundancy measurement
in Sec.~\ref{sec:integration}, and whether a shared pool is actually preferable
is tested directly in Sec.~\ref{sec:controlled}. Table~\ref{Table3} gives the
overall comparison, where UniPrompt-CL improves AvgACC on DR
and remains competitive on skin cancer at substantially lower training cost.

\subsection{Comparison with Continual Learning Methods}

Table~\ref{Table3} reports the continual learning performance and
computational efficiency of UniPrompt-CL against four categories
of baselines: rehearsal-based (DER++), architecture-based
(MoE-Adapters), regularization-based (Online EWC), and
prompt-based CL methods (OS-Prompt, OS-Prompt++, CPS-Prompt,
Coda-Prompt, DualPrompt). Since UniPrompt-CL employs a DINOv2-B backbone, we additionally report results for competing prompt-based methods with DINOv2-B where applicable, enabling a backbone-controlled comparison.

\noindent\textbf{Diabetic Retinopathy.}
UniPrompt-CL achieves the highest AvgACC of 0.844, surpassing the strongest DINOv2-B baseline, Coda-Prompt, by 2.0 percentage points. It also obtains the lowest AvgF (0.079) among all evaluated method--backbone configurations. These results suggest that the proposed unified prompt-pool design improves retained performance beyond the benefit of using a stronger
DINOv2-B backbone.

Both OS-Prompt and UniPrompt-CL preserve the practical advantage of single-ViT
inference, avoiding the two-ViT pipeline required by conventional prompt-based
continual learning methods. Under the same DINOv2-B backbone, UniPrompt-CL
attains higher AvgACC (0.844 vs.\ 0.801) at a lower training cost (44.17 vs.\
66.42 GFLOPs).

\noindent\textbf{Skin Cancer.}
On the skin cancer benchmark, UniPrompt-CL achieves the highest
AvgACC of 0.732, although the margin over the strongest DINOv2-B
baseline, DualPrompt (0.728), is small and lies within the reported
standard deviations. We therefore do not over-interpret the raw
accuracy gap. The main advantage is that UniPrompt-CL maintains
competitive retained accuracy with lower forgetting and substantially
lower compute than DINOv2-B prompt baselines. In particular, compared
with DualPrompt using DINOv2-B, UniPrompt-CL reduces AvgF from 0.079
to 0.049 and GFLOPs from 137.93 to 44.17. Compared with
OS-Prompt++ using DINOv2-B, it also improves AvgACC while reducing
forgetting and computational cost. We also note that DualPrompt with ViT-B
obtains the lowest absolute forgetting on this benchmark (AvgF 0.012), but its
lower AvgACC indicates that minimizing forgetting alone does not yield the best
overall continual-learning utility.

\noindent\textbf{Effect of the backbone.}
A stronger backbone does not uniformly improve every method.
Replacing ViT-B with DINOv2-B raises OS-Prompt++
(0.769$\rightarrow$0.812 on DR) and Coda-Prompt
(0.688$\rightarrow$0.824), but does not help CPS-Prompt,
whose diabetic-retinopathy AvgACC decreases
(0.769$\rightarrow$0.692). This indicates that DINOv2 features are
not universally beneficial across prompt-selection designs. Therefore,
the gains of UniPrompt-CL over the DINOv2-B baselines are better
explained by its prompt-pool organization and expansion strategy than
by backbone choice alone.

\noindent\textbf{Computational efficiency.}
UniPrompt-CL requires only 44.17 GFLOPs, achieving the lowest computational cost among all DINOv2-B configurations. Notably, it is substantially more efficient than OS-Prompt with DINOv2-B, the next lowest-cost DINOv2-B baseline at 66.42 GFLOPs, and CPS-Prompt with DINOv2-B, an efficiency-oriented method that still requires 92.69 GFLOPs. In contrast, Coda-Prompt with DINOv2-B incurs 175.12 GFLOPs—nearly four times the cost of UniPrompt-CL—while delivering lower accuracy and stronger forgetting on both benchmarks. These figures are per-step training cost, including forward and backward passes;
they concern the cost of adapting the model when a new domain arrives, not
inference cost at deployment. Per-dataset final-stage accuracy and F1-score are reported in Sup.~G (Table~8), and a stage-wise forgetting breakdown against OS-Prompt++ is provided in Sup.~H (Table~9).

\begin{table}[t]
\centering
\footnotesize
\setlength{\tabcolsep}{3.5pt}
\renewcommand{\arraystretch}{1.0}
\caption{Unified (U) vs.\ layer-wise (L) prompt pools; only the pool
organization differs. The layer-wise control spreads the same $L_p=500$ prompts
over the five prompted layers, so prompt parameters and unique prompt counts
match. Rows (DR: diabetic retinopathy; Skin: skin cancer) list the two medical
sequences followed by the two natural-image sequences. Means over three seeds; $\Delta=\mathrm{U}-\mathrm{L}$ in points,
with $^\ast$ marking a $95\%$ paired CI excluding zero. These are ablations
within our framework, so absolute values are not comparable to published
DomainNet or OfficeHome results.}
\label{tab:controlled}
\begin{tabular}{lccccc}
\toprule
\multirow{2}{*}{\textbf{Sequence}} & \multirow{2}{*}{\textbf{Stages}} &
\multicolumn{2}{c}{\textbf{AvgACC}$\uparrow$} &
\multicolumn{2}{c}{\textbf{AvgF}$\downarrow$} \\
\cmidrule(lr){3-4} \cmidrule(lr){5-6}
 & & U\;/\;L & $\Delta$ & U\;/\;L & $\Delta$ \\
\midrule
DR         & 3 & 0.844\;/\;0.810 & $+3.44$ & 0.079\;/\;0.172 & $-9.28^\ast$ \\
Skin       & 3 & 0.732\;/\;0.720 & $+1.20$ & 0.049\;/\;0.137 & $-8.75$ \\
OfficeHome & 4 & 0.809\;/\;0.802 & $+0.68$ & 0.114\;/\;0.147 & $-3.29$ \\
DomainNet  & 6 & 0.645\;/\;0.649 & $-0.44$ & 0.506\;/\;0.476 & $+3.02^\ast$ \\
\bottomrule
\end{tabular}
\end{table}

\subsection{Unified vs.\ Layer-wise Prompt Pools}
\label{sec:controlled}

The redundancy analysis in Sec.~\ref{sec:integration} motivates cross-layer
sharing but does not by itself establish that a unified pool is preferable to
layer-wise pools. We therefore vary only the pool organization, holding
everything else fixed, and evaluate the two variants on both medical sequences
and on two natural-image sequences. For the natural-image runs we transfer
UniPrompt-CL unchanged, with no method-specific re-tuning; only
dataset-dependent quantities such as the label space differ. OfficeHome is
presented in the order Art $\rightarrow$ Clipart $\rightarrow$ Product
$\rightarrow$ Real-World, and DomainNet as real $\rightarrow$ clipart $\rightarrow$ painting
$\rightarrow$ sketch $\rightarrow$ infograph $\rightarrow$ quickdraw over all $345$ classes.

Table~\ref{tab:controlled} reports the comparison. On DR the
unified pool improves AvgACC by $3.44$ points and reduces forgetting by $9.28$
points; on skin cancer the accuracy difference falls within the reported
variability but forgetting is again substantially lower. Across the medical
sequences, then, cross-layer sharing acts more consistently on retention than
on per-domain accuracy.

The natural-image sequences show that this benefit is not universal. On
OfficeHome the unified pool remains ahead on both measures, but by a markedly
smaller margin, and the paired confidence intervals include zero. On DomainNet
the ordering reverses: the layer-wise control retains more, lowering AvgF by
$3.02$ points ($95\%$ CI $[2.53, 3.50]$, $3/3$ seeds); the same holds for final average accuracy (FAA), which is 2.49 points higher
for the layer-wise control (95\% CI [2.03, 2.95], 3/3 seeds). Read down the $\Delta$ column
for AvgF, the advantage of sharing shrinks from $-9.28$ to $-8.75$ to $-3.29$
and finally changes sign at $+3.02$ across the four sequences, ordered from
acquisition shifts, through changes of visual style, to the six-domain
DomainNet stream.

Two features of the DomainNet result make it informative beyond the sign
change. First, the two variants learn each new domain similarly---the diagonal
entries differ by at most $0.44$ points at every stage---so the difference is
concentrated in retention rather than in initial acquisition, and it arises
under an identical prompt budget. Second, the difference does not appear in
AvgACC, whose confidence interval includes zero. This is consistent with the
metric definitions: AvgACC mixes immediate adaptation with retained performance
across the whole trajectory and therefore dilutes an effect concentrated in
final retention.

What the experiment establishes is that the benefit of sharing is
shift-dependent rather than universal: largest on the medical acquisition
shifts we target, smaller on OfficeHome, and reversed on the six-domain
DomainNet stream. This pattern is consistent with our hypothesis that
cross-layer sharing helps when the required adaptation is reusable across
representation depths and hurts when a shift demands distinct adaptation at
different depths, though we do not measure that property directly and state
the claim accordingly. DomainNet is
also a stress test in absolute terms: after six stages both variants retain
under $0.35$ FAA, so it probes the boundary of the design rather than a regime
in which either variant performs well.

\subsection{Ablation Study}
\label{sec:ablation}

\noindent\textbf{Impact of the Proposed Loss Term.}
Fig.~\ref{fig:loss_ablation} evaluates the surrogate selection loss
$\mathcal{L}_s$ and its weight $\lambda$, reported as mean$\pm$std over three
seeds. FAA denotes the final average accuracy across all stages and FAF the
final average F1-score. Removing $\mathcal{L}_s$ ($\lambda=0$) is the worst
setting on both measures, and $\lambda=0.001$ is the best on both: FAA
improves from $0.720\pm0.031$ to $0.750\pm0.022$ and FAF from
$0.674\pm0.028$ to $0.701\pm0.014$. Performance degrades again at
$\lambda=0.01$. We therefore adopt $\lambda=0.001$. With three seeds the differences at individual settings lie
within the reported standard deviations, so we do not claim a statistically
resolved optimum; the evidence for $\mathcal{L}_s$ is rather that both
measures rank the settings identically, with FAF favouring $\lambda=0.001$
over $\lambda=0$ in all three seeds. Stage-1 performance provides a control: there
$\mathcal{I}_{\psi}=\emptyset$ and $\mathcal{L}_s$ is identically zero, so
$\lambda$ cannot act, and the spread we still observe across $\lambda$ within a
seed (s.d.\ $0.77$ points) measures the run-to-run variability of our setup.
The effect of $\mathcal{L}_s$ on FAA, $3.08$ points, is four times that figure.

\noindent\textbf{Effect of Prompt Expansion Ratio.}
\label{sec:prompt_ratio}
Fig.~\ref{fig:prompt_expansion_ratio} sweeps the expansion ratio
$\rho \in \{0, 10, 20, 30, 50, 100\}\%$ on APTOS $\rightarrow$ DDR
$\rightarrow$ DRD over three seeds, extending the range in both directions so
that the no-expansion case and a full doubling of the pool are both covered.

Expansion itself is what matters most. Removing it entirely ($\rho=0$) is the
worst setting on all four measures: relative to $\rho=20\%$ it loses $6.84$
points of Acc-based average performance (95\% CI $[5.36, 8.32]$, $3/3$ seeds)
and $13.97$ points of FAA (95\% CI $[3.99, 23.94]$, $3/3$ seeds), while
forgetting rises from $0.105$ to $0.283$. This is the comparison we designate
in advance; the remaining settings are reported without further testing.

A larger pool, however, does not buy proportionally more. Beyond $\rho=20\%$
average performance is flat within seed variability ($0.812$, $0.818$ and
$0.810$ at $30\%$, $50\%$ and $100\%$, against a run-to-run standard deviation
of $0.93$ points), whereas forgetting increases monotonically ($0.105
\rightarrow 0.159 \rightarrow 0.193 \rightarrow 0.200$). Quintupling the number
of added prompts therefore costs retention without returning accuracy. The
extreme settings are also the least stable: the seed standard deviation of AvgF
is $0.061$ at $\rho=0$ and $0.109$ at $\rho=100\%$, against $0.034$ at
$\rho=20\%$. The sweep thus identifies a capacity--retention trade-off rather
than a monotone benefit of a larger pool, and we adopt $\rho=20\%$, which
attains the best value on all four measures and the smallest variance among the
ratios we evaluate.

\begin{figure}[t]
\centering
\includegraphics[width=0.45\columnwidth]{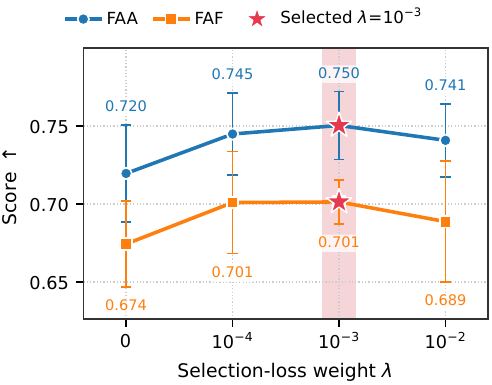}
\caption{Ablation analysis of the surrogate selection loss $\mathcal{L}_s$ and
its weight $\lambda$ on APTOS $\rightarrow$ DDR $\rightarrow$ DRD. All values are
computed on the validation split, not on the test split. FAA and FAF
denote the final average accuracy and final average F1-score across all stages,
respectively. Markers show the mean over three seeds and error bars one standard
deviation. The selected setting, $\lambda=0.001$, attains the best value on both
measures; performance declines again at larger $\lambda$.}
\label{fig:loss_ablation}
\end{figure}

\begin{figure}[t]
\centering
\includegraphics[width=\columnwidth]{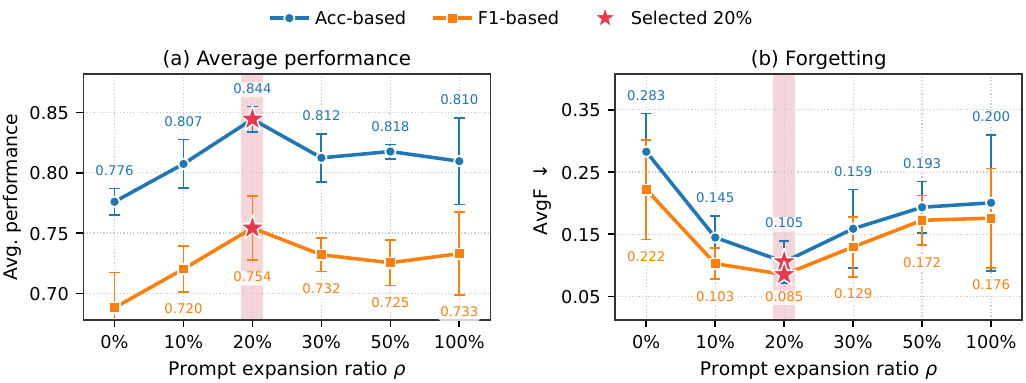}
\caption{Prompt expansion ratio sensitivity on the APTOS $\rightarrow$ DDR
$\rightarrow$ DRD sequence. All values are computed on the validation split of
each dataset, not on the test split. Markers show the mean over three seeds and error
bars one standard deviation; the horizontal axis is categorical, with the six
evaluated ratios at equal spacing. The highlighted $20\%$ ratio attains the
best value on all four measures. Beyond it, average performance is flat while
forgetting continues to grow.}

\label{fig:prompt_expansion_ratio}
\end{figure}

\section{Conclusion}
\label{sec:Conclusion}
We presented UniPrompt-CL, a rehearsal-free prompt-based framework for
domain-incremental medical image classification that replaces the layer-wise
independent prompt pools of prior work with a single pool shared across
transformer layers, expanded by a few prompts per stage and regularized so that
newly added prompts are exercised early. The design keeps the single-backbone,
single-inference property of OS-Prompt.

On DR it improves AvgACC over the strongest
backbone-matched baseline and reaches comparable accuracy on skin cancer, in
both cases with the lowest forgetting and the lowest training cost among the
DINOv2-B configurations we evaluate. Since a stronger backbone does not help
every method uniformly, these gains follow from how prompt capacity is
organized rather than from the backbone alone.

A comparison that varies only the pool organization both supports this and
bounds it. The advantage of sharing is concentrated in retention rather than in
learning each new domain; it is largest on the medical benchmarks, smaller on
OfficeHome, and reverses on a more heterogeneous six-stage DomainNet stream,
where layer-wise pools retain more than a shared pool of the same size. We therefore
do not claim the design generalizes to shifts of that kind, and predicting in
advance which regime a stream falls into remains open. A second limitation is the fixed expansion ratio $\rho$; the sweep in
Sec.~\ref{sec:ablation} shows that what a larger pool costs is retention rather
than capacity, and accounting for that trade-off is a natural next step.
Finally, our evaluation reports accuracy and macro-F1 but not per-class
sensitivity; for a graded disease the sensitivity on the severe minority grade
is the operative quantity, and we leave that breakdown to future work.

\section*{Acknowledgment}
This work was supported by the BK21 FOUR Project and by the National Research Foundation of Korea (NRF) grant funded by the Korean government (MSIT) (RS-2026-25589675). 

\clearpage
\end{cbunit}

\begin{cbunit}
\clearpage
\appendix

\twocolumn[{%
  \centering
  \Large
  \textbf{\thetitle}\\
  \vspace{0.5em}Supplementary Material \\
  \vspace{1.0em}
  \normalsize
  \begin{minipage}{\textwidth}
  \section*{Overview}
  This supplementary material is organized in the order in which its sections are
  cited in the main paper. Each row below names a section and states what it adds
  to the main text.
  \end{minipage}

  \begin{center}
  \small
  \setlength{\tabcolsep}{6pt}
  \renewcommand{\arraystretch}{1.2}
  \begin{tabular}{@{}c p{0.33\textwidth} p{0.56\textwidth}@{}}
  \toprule
  \textbf{Sec.} & \textbf{Title} & \textbf{What it adds to the main paper} \tabularnewline
  \midrule
  \textbf{\ref{sup:osprompt}} &
    \raggedright Background: OS-Prompt/OS-Prompt++ and Differences from UniPrompt-CL &
    \raggedright How OS-Prompt and OS-Prompt++ form their prompt queries and pools, and
    where UniPrompt-CL departs from them. \tabularnewline
  \textbf{\ref{sec:sup_same_index_redundancy}} &
    \raggedright Cross-Layer Same-Index Redundancy Analysis of OS-Prompt &
    \raggedright Quantitative evidence that OS-Prompt's layer-wise pools are redundant
    across layers at the same index, which motivates the unified pool. \tabularnewline
  \textbf{\ref{sup:data}} &
    \raggedright Medical Dataset Statistics and Backbone Choice &
    \raggedright Class distributions of the DR and skin cancer datasets, and the
    rationale for selecting DINOv2-B. \tabularnewline
  \textbf{\ref{A_E}} &
    \raggedright UniPrompt-CL Training Procedure &
    \raggedright Pseudocode for training with fixed per-stage prompt expansion. \tabularnewline
  \textbf{\ref{A_B}} &
    \raggedright Implementation and Reproducibility Details &
    \raggedright Preprocessing, experimental settings, and baseline hyperparameters. \tabularnewline
  \textbf{\ref{A_C}} &
    \raggedright Continual Learning Metrics &
    \raggedright Definitions of AvgACC, BWT, AvgF, FAA, and FAF, and the auxiliary
    compute-adjusted metric CARA, whose values and sensitivity analysis are reported
    only here. \tabularnewline
  \textbf{\ref{sec:appendix_perdataset}} &
    \raggedright Per-Dataset Final-Stage Results &
    \raggedright Final-stage accuracy and F1-score on each dataset for all compared
    methods. \tabularnewline
  \textbf{\ref{sec:appendix_stagewise}} &
    \raggedright Additional Stage-Wise Forgetting Analysis &
    \raggedright Stage-by-stage comparison of retained performance between OS-Prompt++
    and UniPrompt-CL. \tabularnewline
  \textbf{\ref{A_A}} &
    \raggedright GenAI Usage Disclosure &
    \raggedright How generative AI tools were used in preparing the manuscript. \tabularnewline
  \bottomrule
  \end{tabular}
  \end{center}
}]

\section{Background: OS-Prompt/OS-Prompt++ and Differences from UniPrompt-CL}
\label{sup:osprompt}

Here, we detail the foundational OS-Prompt framework and its extension
OS-Prompt++~\cite{kim2024one}, and clarify how our method differs from them.
Both OS-Prompt and OS-Prompt++ aim to overcome the high computational cost of
traditional prompt-based continual learning (PCL), which stems from an additional query
function, such as a separate ViT, used to generate prompt queries. OS-Prompt removes
this auxiliary query function and instead uses the layer-wise [CLS] token from the
backbone as the prompt query:
\begin{equation}
q_l = x_{l}[\mathrm{CLS}],
\end{equation}
which corresponds to Eq.~1 in the main text.

To retrieve prompts from the pool, OS-Prompt follows the soft-matching formulation of
CodaPrompt~\cite{smith2023coda}, based on cosine-similarity weighted aggregation. At
layer $l$, the prompt pool is defined with layer-specific prompt keys and values:
\begin{equation}
P_l = \{(k_l^m, p_l^m)\}_{m=1}^{L_p},
\end{equation}
where $k_l^m \in \mathbb{R}^{1 \times D}$ is the key of the $m$-th prompt and
$p_l^m \in \mathbb{R}^{1 \times D}$ is its corresponding prompt value. The prompt key
matrix $K_l$ is formed by stacking the $L_p$ key vectors, yielding
$K_l \in \mathbb{R}^{L_p \times D}$. Given the layer-wise query $q_l$, the selected
prompt is computed as
\begin{equation}
\hat{\phi}_l =
\sum_{m=1}^{L_p}
\gamma(q_l, k_l^m) p_l^m,
\quad
\hat{\phi}_l \in \mathbb{R}^{1 \times D},
\label{eq:osprompt-select}
\end{equation}
where $\gamma(\cdot)$ denotes the cosine-similarity-based prompt weight used in OS-Prompt.

\begin{figure*}[t]
\centering
\includegraphics[width=\linewidth]{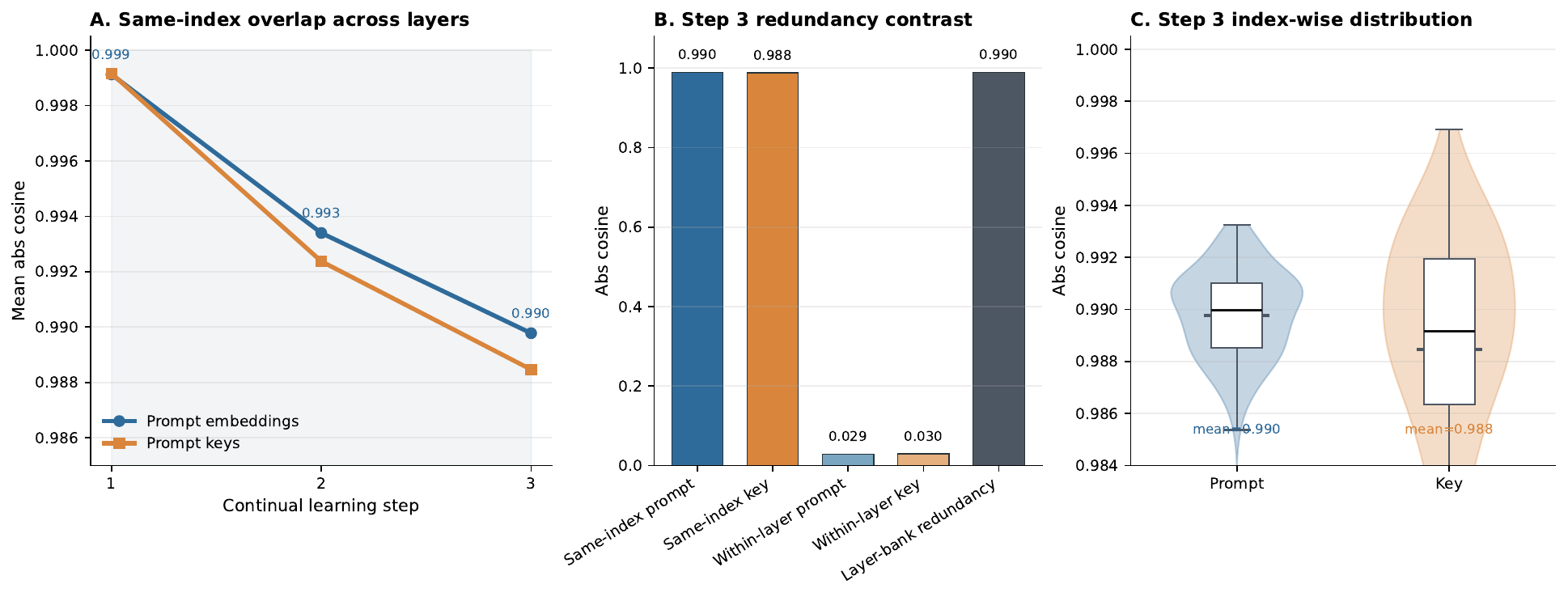}
\caption{
Cross-layer same-index redundancy analysis of OS-Prompt.
(A) Prompts and keys with the same index across different prompted layers exhibit consistently high absolute cosine similarity across continual learning stages.
(B) At the final stage, same-index cross-layer similarity is close to one, while
within-layer prompt/key similarity remains close to zero.
(C) The index-wise distribution confirms that high same-index redundancy is
observed for most prompt indices, rather than a small subset.
}
\label{fig:sup_os_prompt_redundancy_diagnostics}
\end{figure*}

\paragraph{Prompt-pool structure.}
The key difference between OS-Prompt and our method lies in the prompt-pool structure.
OS-Prompt uses \emph{layer-wise independent prompt pools}, where each prompted layer has
its own set of keys and values $\{k_l^m, p_l^m\}_{m=1}^{L_p}$. In contrast, our method
uses a \emph{unified prompt pool} shared across layers, as described in
Eq.~2. This design is motivated by the observation that the medical
datasets we study are acquired under relatively standardized protocols, where domain
shifts are subtle and fine-grained. Under such conditions, maintaining independent prompt
pools at each layer may introduce redundant prompt capacity across layers. The unified
prompt pool instead encourages prompt sharing across layers and promotes more efficient
use of limited prompt parameters.

OS-Prompt++ further extends OS-Prompt by introducing Query-Pool Regularization (QR)
during training. Since OS-Prompt uses an intermediate [CLS] token as the prompt query,
its query representation can be weaker than the final-layer [CLS] representation used
in conventional two-stage PCL. To compensate for this, OS-Prompt++ forwards the input
through an additional reference ViT during training and extracts a reference query from
the final-layer [CLS] token. It then aligns the query-pool similarity matrix from the
backbone query with that from the reference query:
\begin{equation}
A^l_{\mathrm{query}} =
\mathrm{Softmax}
\left(
\frac{K_l q_l^\top}{\|K_l\|_2 \|q_l\|_2}
\right),
\end{equation}
\begin{equation}
A^l_{\mathrm{ref}} =
\mathrm{Softmax}
\left(
\frac{K_l r^\top}{\|K_l\|_2 \|r\|_2}
\right),
\end{equation}
\begin{equation}
\mathcal{L}_{\mathrm{QR}}
=
\sum_l
\left\|
A^l_{\mathrm{query}} -
A^l_{\mathrm{ref}}
\right\|_2^2 .
\end{equation}
This QR loss improves the representational quality of the intermediate-layer query.
However, it requires an additional reference ViT forward pass during training. Therefore,
OS-Prompt++ is efficient at inference time, but it does not provide a strictly
single-backbone training procedure.

Our method differs from OS-Prompt++ in both objective and mechanism. We do not introduce
a reference ViT or align intermediate queries with final-layer reference queries.
Instead, we retain the single-backbone query mechanism of OS-Prompt and focus on how the
prompt pool itself should be organized and expanded under domain-incremental
shifts. Specifically, our method replaces the layer-wise independent prompt pools with
a unified prompt pool and expands this pool by adding only a small number of new prompts
when a new domain arrives. In addition, we introduce a surrogate selection loss applied
only to the newly added prompt subset, which encourages those prompts to be
retrieved during the early phase of domain expansion.

In summary, OS-Prompt mainly addresses the computational inefficiency of query
generation, and OS-Prompt++ improves the representation of intermediate queries through
training-time reference-ViT regularization. In contrast, our method addresses
prompt-pool organization and expansion under domain-incremental shifts. The
proposed design preserves the single-backbone inference advantage of OS-Prompt while
improving prompt utilization through unified prompt sharing, lightweight prompt
expansion, and selection regularization.

\section{Cross-Layer Same-Index Redundancy Analysis of OS-Prompt}
\label{sec:sup_same_index_redundancy}

Sec.~3.2 of the main paper reports that the layer-wise prompt
pools of OS-Prompt carry substantial overlap across layers. Those numbers come
from the quantitative same-index redundancy analysis given here, which uses a
trained OS-Prompt checkpoint. The goal is
not to show that the entire prompt pool collapses, but to examine whether prompts
assigned to the same index across different layers learn distinct layer-specific
representations.

Let $p_l^m$ and $k_l^m$ denote the prompt value and prompt key of OS-Prompt at
prompted layer $l \in \mathcal{I}$ and prompt index $m \in \{1,\ldots,L_p\}$,
respectively. For each prompt index $m$, we compute the average cross-layer
same-index prompt similarity as
\begin{equation}
R_{\mathrm{prompt}}(m) =
\frac{2}{|\mathcal{I}|\,(|\mathcal{I}|-1)}
\sum_{\substack{l,l'\in\mathcal{I}\\ l<l'}}
\left|
\cos\!\left(p_l^m,\, p_{l'}^m\right)
\right|,
\end{equation}
and define the corresponding same-index key similarity as
\begin{equation}
R_{\mathrm{key}}(m) =
\frac{2}{|\mathcal{I}|\,(|\mathcal{I}|-1)}
\sum_{\substack{l,l'\in\mathcal{I}\\ l<l'}}
\left|
\cos\!\left(k_l^m,\, k_{l'}^m\right)
\right|.
\end{equation}
These quantities measure whether the $m$-th prompt or key in one layer points in
a similar direction to the $m$-th prompt or key in other layers.

For comparison, we also measure within-layer prompt diversity by computing the
average absolute cosine similarity between different prompt indices within the
same layer:
\begin{equation}
D_{\mathrm{prompt}} =
\frac{1}{|\mathcal{I}|}
\sum_{l\in\mathcal{I}}
\frac{2}{L_p(L_p-1)}
\sum_{1\leq m<n\leq L_p}
\left|
\cos\!\left(p_l^m,\, p_l^n\right)
\right|,
\end{equation}
with $D_{\mathrm{key}}$ defined analogously for prompt keys. This comparison
separates two distinct phenomena: diversity among different prompts within a
layer, and redundancy of the same prompt index across layers.

As shown in Sup.~Fig.~\ref{fig:sup_os_prompt_redundancy_diagnostics}(A), prompts
with the same index across different layers exhibit very high absolute cosine
similarity. Panel~(B) contrasts this same-index redundancy with within-layer
diversity: while same-index prompt and key similarities across layers are close
to one, the within-layer similarities remain close to zero. OS-Prompt therefore
does not collapse its prompt pool within each layer---each layer keeps its
prompts internally diverse---yet prompts sharing the same index across layers
are highly redundant. Panel~(C) reports the index-wise distribution of
same-index overlap at the final stage, where most indices consistently show high
cross-layer similarity. Together, these results indicate that the layer-wise
independent prompt pools of OS-Prompt can carry substantial cross-layer
same-index representational redundancy, motivating our unified prompt-pool
design.

\section{Medical Dataset Statistics and Backbone Choice}
\label{sup:data}

As shown in Sup.~Tables~\ref{Table8} and~\ref{Table9}, medical datasets often exhibit
severe long-tailed class distributions, which exacerbate catastrophic forgetting as
training stages progress. To mitigate this, we adopt the DINOv2 backbone~\cite{oquab2023dinov2},
selecting DINOv2-base (86.6M parameters) for its balanced trade-off between
representational capacity and computational efficiency. The model-size comparison is
reported in Sup.~Table~\ref{Table12}. Combined with our prompt-based design and
regularization, this backbone effectively mitigates both data imbalance and catastrophic
forgetting in medical CL scenarios.

\begin{table}[ht]
\centering
\caption{Distribution of diabetic retinopathy datasets used for training.}
\label{Table8}
\begin{tabular}{lrrr}
\hline
\textbf{Dataset} & \textbf{Normal (0)} & \textbf{NPDR (1)} & \textbf{PDR (2)} \\
\hline
APTOS~\cite{aptos2019-blindness-detection} & 1805 & 1562 & 295 \\
DDR~\cite{LI2019}       & 6266 & 5343 & 913 \\
DRD~\cite{diabetic-retinopathy-detection}   & 5000 & 8608 & 708 \\
\hline
\end{tabular}
\end{table}

\begin{table}[ht]
\centering
\caption{Distribution of skin cancer datasets used for external validation.}
\label{Table9}
\begin{tabular}{lrr}
\hline
\textbf{Dataset} & \textbf{Benign} & \textbf{Malignant} \\
\hline
ISIC~\cite{rotemberg2021patient} & 584 & 584 \\
HAM~\cite{codella2019skin}       & 1113 & 1113 \\
DERM7~\cite{kawahara2018seven}   & 252 & 252 \\
\hline
\end{tabular}
\end{table}

\begin{table}[ht]
\centering
\caption{Parameter counts for facebook/dinov2 models.}
\label{Table12}
\begin{tabular}{lc}
\toprule
Model & Parameters \\
\midrule
facebook/dinov2-small  & 22.1\,M  \\
facebook/dinov2-base   & 86.6\,M  \\
facebook/dinov2-large  & 304\,M   \\
facebook/dinov2-giant  & 1.14\,B  \\
\bottomrule
\end{tabular}
\end{table}

In general, models with a larger number of parameters can extract richer features and
thus achieve better performance. However, as model size increases, the parameter count
grows more than twofold at each step, leading to significantly higher complexity. For
this reason, we adopted the moderately sized DINOv2-base model. These models were
employed using the Hugging Face library.

\begin{algorithm}[t]
\caption{UniPrompt-CL Training Procedure with Fixed Per-Stage Prompt Expansion}
\label{alg:uniprompt_cl}
\begin{algorithmic}[1]
\Require Dataset sequence $\{\mathcal{D}_n\}_{n=1}^{N}$, frozen backbone $\mathcal{M}$, classifier head $h$, initial unified prompt pool $(K,P)$, expansion ratio $\rho$, regularization weight $\lambda$
\Ensure Trained classifier head $h$ and unified prompt pool $(K,P)$

\State Freeze the backbone $\mathcal{M}$
\State Initialize the unified prompt pool $(K,P)$
\State Set the number of newly added prompts as $\psi=\lfloor \rho L_p \rfloor$

\For{each continual learning stage $n=1,\dots,N$}
    \If{$n>1$}
        \State Freeze prompts learned from previous stages
        \State Add $\psi$ newly initialized prompt keys and values to $(K,P)$
        \State Define $\mathcal{I}_{\psi}$ as the index set of newly added prompts
    \Else
        \State Set $\mathcal{I}_{\psi}=\emptyset$ and $\mathcal{L}_s=0$
    \EndIf

    \For{each mini-batch $(X,Y)\sim \mathcal{D}_n$}
        \For{each prompted layer $l\in\mathcal{I}$}
            \State Obtain the layer-wise query $q_l$ from the [CLS] token
            \State Retrieve the prompt $\phi_l$ from the unified prompt pool using
                   similarity-weighted aggregation (Eq.~2)
            \State Fuse $\phi_l$ with the layer representation
        \EndFor

        \State Compute prediction $\hat{Y}$ from the prompt-augmented backbone output
        \State Compute classification loss $\mathcal{L}_{\mathrm{CE}}$

        \If{$n>1$}
            \State Compute the surrogate prompt-selection loss $\mathcal{L}_s$ over $\mathcal{I}_{\psi}$
        \EndIf

        \State Update trainable parameters using
        \[
        \mathcal{L}_{\mathrm{total}}
        =
        \mathcal{L}_{\mathrm{CE}}
        +
        \lambda \mathcal{L}_s
        \]
    \EndFor
\EndFor

\State \Return trained classifier head $h$ and unified prompt pool $(K,P)$
\end{algorithmic}
\end{algorithm}
\section{UniPrompt-CL Training Procedure}
\label{A_E}

We present Sup.~Algorithm~\ref{alg:uniprompt_cl} to clarify the training process and the
behavior of UniPrompt-CL under fixed per-stage prompt expansion. Here $\mathcal{I}$
denotes the set of prompted layers, consistent with the main text.

\section{Implementation and Reproducibility Details}
\label{A_B}

To ensure reproducibility and completeness, we summarize the preprocessing procedure,
external validation setup, implementation environment, and baseline hyperparameters for
both the main diabetic retinopathy (DR) study and the supplementary skin-cancer experiments.

\textbf{Main datasets (Diabetic Retinopathy).}
Our primary experiments use three publicly available DR datasets:
APTOS, DDR, and DRD. Since some benchmarks do not provide ground-truth labels for the
official test sets, we follow prior work~\cite{kobat2022automated} to preprocess the
data and construct consistent train/validation/test splits. Following the International
Clinical DR (ICDR) severity scale, mild/moderate/severe stages
correspond to nonproliferative DR, while PDR is a distinct advanced
proliferative stage; accordingly, and as in~\cite{kobat2022automated}, we remap the
original five DR grades (normal, mild, moderate, severe, proliferative) into three
clinically grounded classes: normal, NPDR, and PDR, by merging mild/moderate/severe into
NPDR while keeping normal and PDR unchanged. This re-mapping is intended to yield a
consistent, clinically meaningful label space shared across the three datasets, rather
than to simplify the task. In our continual learning setting, the training stream is
denoted as
\[
D = \{D_1, D_2, \ldots, D_n\}, \quad D_n = \{X_{n,b}, Y_{n,b}\},
\]
where \(n\) is the stage index, \(X_{n,b}\) is an input image, \(Y_{n,b}\) is the
corresponding label, and \(b\) indexes samples within a batch of size \(B\).

\textbf{Supplementary external datasets (Skin Cancer).}
To assess generalization beyond DR, we conduct small-scale external experiments under
the same domain-incremental protocol, using three skin-cancer datasets in the order
ISIC $\rightarrow$ HAM $\rightarrow$ DERM7, yielding three stages. Following prior work
on ISIC lesion grouping~\cite{cassidy2022analysis}, we remap the original diagnostic
categories into a clinically motivated binary setting: lesions requiring urgent clinical
intervention are grouped as \textit{malignant}, while others are grouped as benign.
Because ISIC, HAM, and DERM7 differ substantially in size, diagnostic taxonomy, and
benign/malignant priors, using all images directly could confound continual-learning
behavior with dataset scale and class imbalance. We therefore construct class-balanced
subsets within each domain so that the external study isolates cross-domain transfer
rather than dataset-scale effects; this is a controlled external validation, not a
full-scale skin-lesion benchmark. All datasets are split 8:1:1 into
train/validation/test sets. All other settings, including model, training schedule,
hyperparameters, and metrics, are identical to the DR study; only the datasets change,
and we set the batch size to 16 due to the smaller dataset size. Representative examples of the skin cancer datasets used for external validation are shown in Figure~\ref{Figure4}.

\begin{figure}[ht!]
\centering
\includegraphics[width=\columnwidth]{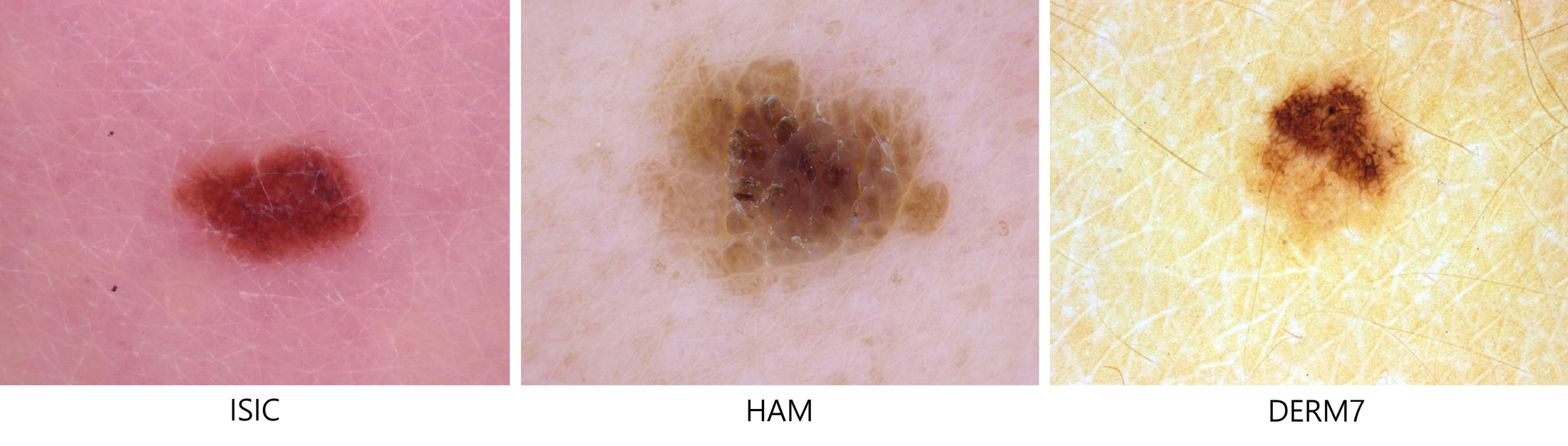}
\caption{Representative examples from the three skin cancer datasets used for the controlled external validation: ISIC~\cite{rotemberg2021patient}, HAM~\cite{codella2019skin}, and DERM7~\cite{kawahara2018seven}. The samples illustrate the diversity in lesion appearance and acquisition conditions across datasets.}
\label{Figure4}
\end{figure}

\begin{table}[h!]
\centering
\caption{Experiment settings.}
\label{Table10}
\begin{tabular}{ll}
\toprule
\textbf{Item} & \textbf{Value} \\
\midrule
GPU                   & 2 $\times$ V100 \\
RAM                   & 256GB \\
Learning Rate         & $0.001$ \\
Scheduler             & Cosine Scheduler \\
Number of Prompts ($L_p$)    & 500 \\
Prompt Dimension ($D$)       & 768 \\
Optimizer             & AdamW \\
Early Stop Patience   & 5 \\
Epochs                & 100 \\
Batch Size            & 64 \\
Image Size            & 224 $\times$ 224 \\
\bottomrule
\end{tabular}
\end{table}

\begin{table}[t]
\centering
\caption{Key hyperparameters for prompt-based continual learning baselines.}
\label{Table11}
\resizebox{\linewidth}{!}{%
\begin{tabular}{lll}
\toprule
Method & Hyperparameter & Value \\
\midrule
\multirow{3}{*}{CODA-Prompt}
 & Prompt pool size ($\texttt{coda\_pool\_size}$) & 100 \\
 & Prompt length ($\texttt{coda\_prompt\_length}$) & 4 \\
 & Ortho. regularization coeff. ($\texttt{coda\_ortho\_mu}$) & 0.1 \\
\midrule
\multirow{7}{*}{DualPrompt}
 & Prompt length ($\texttt{dual\_prompt\_length}$) & 4 \\
 & Prompt pool size ($\texttt{dual\_pool\_size}$) & 100 \\
 & Top-$k$ prompts ($\texttt{dual\_top\_k}$) & 5 \\
 & G-prompt length ($\texttt{dual\_g\_prompt\_length}$) & 4 \\
 & G-prompt layers ($\texttt{dual\_g\_prompt\_layers}$) & [0, 1] \\
 & E-prompt layers ($\texttt{dual\_e\_prompt\_layers}$) & [2, 3, 4] \\
 & Head type ($\texttt{dual\_head\_type}$) & token \\
\midrule
\multirow{3}{*}{DER++}
 & ALPHA (\texttt{larger values emphasize preserving past knowledge}) & 0.5 \\
 & BETA (\texttt{larger values emphasize relearning past tasks}) & 0.5 \\
 & BUFFER\_SIZE (\texttt{number of stored past samples}) & 2000 \\
\midrule
\multirow{3}{*}{Online EWC}
 & RP\_SIZE (\texttt{dimension of the random projection matrix}) & 10000 \\
 & OPTIM\_MOM (\texttt{SGD momentum}) & 0.9 \\
 & OPTIM\_WD ($\ell_2$ \texttt{regularization coefficient}) & 0.0005 \\
\midrule
\multirow{3}{*}{MoE Adapters}
 & PROMPT\_TEMPLATE (\texttt{CLIP text prompt template}) & ``a photo of \{\}.'' \\
 & CLIP\_BACKBONE (\texttt{CLIP model backbone}) & ViT-B/16 \\
 & VIRTUAL\_BS\_N (\texttt{virtual batch size iterations}) & 1 \\
\midrule
\multirow{2}{*}{OS-Prompt / OS-Prompt++}
 & Prompt dimension & 768 \\
 & Pool size & 150 \\
 \midrule
\multirow{7}{*}{CPS-Prompt}
 & Prompt pool size ($\texttt{cps\_pool\_size}$) & 100 \\
 & Prompt length ($\texttt{cps\_prompt\_len}$) & 8 \\
 & Phase ratio ($\texttt{cps\_phase\_ratio}$) & 0.8 \\
 & Patch reduction ratio ($\texttt{cps\_reduction\_ratio}$) & 0.5 \\
 & Sampling strategy ($\texttt{cps\_sampling}$) & critical\_score \\
\bottomrule
\end{tabular}}
\end{table}

\textbf{Implementation details.}
Our implementation uses Python 3.8 and PyTorch \textbf{2.4.1+cu118}. All experiments were
conducted in the controlled environment summarized in Sup.~Table~\ref{Table10} and
Sup.~Table~\ref{Table11}. We used two NVIDIA V100 GPUs for training efficiency; however,
the full GPU memory capacity was not required, since PCL methods update only additional
learnable parameters while keeping most of the backbone frozen. In practice, training
with a single V100 required approximately 8\,GB VRAM, supporting reproducibility on
modest hardware.

\textbf{Computational cost measurement.}
The GFLOPs in Table~1 of the main paper are training-time costs
measured under the same protocol for all methods. For each configuration we
count the floating-point operations of one training step on a single
$224 \times 224$ input, comprising the forward pass, the cross-entropy loss, and
the backward pass. For methods that keep the backbone frozen, no parameter
gradients are computed for the backbone weights, while the backward computation
required to propagate gradients through the backbone to the trainable prompts,
adapters, and classifier parameters is included; methods that update the
backbone incur the corresponding backward cost. Because GFLOPs are a
deterministic function of the input shape and the method, a single value is
reported per configuration, and the cost does not depend on the benchmark. The
same procedure is applied to the ViT-B/16 and DINOv2-B configurations.

\section{Continual Learning Metrics}
\label{A_C}

\begin{equation}
\mathrm{AvgACC}
= \frac{1}{T}\sum_{i=1}^{T}\left(\frac{1}{i}\sum_{j=1}^{i} R_{i,j}\right).
\end{equation}

\textbf{Symbols:}
$T$ denotes the total number of stages, and $R_{i,j}$ denotes the performance on task
$j$ after training up to stage $i$.

\textbf{Explanation:}
This metric measures the average performance trajectory throughout the entire training
process. After learning each stage $i$, it computes the average performance over all
stages learned so far and then averages this value across all $T$ stages.

\textbf{Meaning:}
AvgACC provides a comprehensive view of how stably the model maintains its performance
over time. Unlike metrics that only evaluate the final result, it reflects performance
consistency across the whole learning process.

\begin{equation}
\mathrm{BWT}
= \frac{1}{T-1}\sum_{j=1}^{T-1}\bigl(R_{T,j} - R_{j,j}\bigr).
\end{equation}

\textbf{Symbols:}
$R_{j,j}$ denotes the performance on task $j$ immediately after it was first learned,
and $R_{T,j}$ denotes the final performance on task $j$ after completing all $T$ stages.

\textbf{Explanation:}
BWT measures how the performance on past tasks changes after learning new tasks.

\textbf{Meaning:}
If BWT is close to zero, new tasks do not significantly affect old tasks. If BWT is
negative, forgetting occurs. If BWT is positive, new tasks improve the performance on
old tasks, indicating positive backward transfer.

\begin{equation}
\mathrm{AvgF}
= \frac{1}{T-1}\sum_{j=1}^{T-1}\left(\max_{i \in \{j,\dots,T\}} R_{i,j} - R_{T,j}\right).
\end{equation}

\textbf{Symbols:}
$\max_{i \in \{j,\dots,T\}} R_{i,j}$ denotes the best performance achieved on task $j$
during the entire training process.

\textbf{Explanation:}
Average Forgetting (AvgF) measures how much performance is lost on previously learned
tasks after completing all training stages. It compares the best performance achieved on
each task during training with its final performance.

\textbf{Meaning:}
A higher AvgF value indicates greater forgetting, while an AvgF close to zero implies
strong retention of previously learned knowledge.

\textbf{Relation between BWT and AvgF.}
The two definitions differ in their reference point: BWT compares the final
performance on a domain with the performance recorded immediately after that
domain was learned, whereas AvgF compares it with the best performance recorded
at any later point. They coincide up to sign exactly when
$\max_{i\in\{j,\dots,T\}} R_{i,j} = R_{j,j}$, i.e.\ when every domain attains
its maximum immediately after being learned. Sup.~Table~\ref{Table_stagewise}
shows that this holds for accuracy on the two medical sequences, so
$\mathrm{BWT}=-\mathrm{AvgF}$ in every row of Table~1. The equality
is a property of those particular sequences rather than of the metrics, and it
does not hold in general: a small positive backward transfer appears in
F1-score for individual medical runs, and in accuracy for one of the
natural-image runs in Sec.~5.3. We therefore report AvgF only
in the main paper and retain both definitions here.

\begin{table*}[ht]
\centering
\caption{Final Accuracy (Acc) and F1-Score (F1) Results After the Final Stage and Performance Comparison with Other PCL Models. The symbol $\dagger$ denotes our proposed model. PCL denotes prompt-based continual learning; Arch-CL refers to architecture-based continual learning; Reh-CL means rehearsal-based continual learning; and Reg-CL indicates regularization-based continual learning. (\textbf{bold} indicates the highest performance;)}
\label{Table1}
\small
\setlength{\tabcolsep}{4.2pt}
\renewcommand{\arraystretch}{0.95}
\resizebox{\textwidth}{!}{
\begin{tabular}{llcccccc}
\toprule
\textbf{CL-Type (Ref)} & \textbf{Model} & \multicolumn{2}{c}{\textbf{APTOS}} & \multicolumn{2}{c}{\textbf{DDR}} & \multicolumn{2}{c}{\textbf{DRD}} \\
\cmidrule(r){3-4} \cmidrule(r){5-6} \cmidrule(r){7-8}
& & Acc & F1 & Acc & F1 & Acc & F1 \\
\midrule
PCL (CVPR2026)   & CPS-Prompt~\cite{lim2026critical}
& 0.676$\pm$0.046 & 0.651$\pm$0.041
& 0.694$\pm$0.011 & 0.660$\pm$0.018
& 0.647$\pm$0.010 & 0.579$\pm$0.006 \\
PCL (ECCV2024)   & OS-Prompt~\cite{kim2024one}
& 0.687$\pm$0.011 & 0.637$\pm$0.042
& 0.693$\pm$0.013 & 0.648$\pm$0.043
& 0.619$\pm$0.002 & 0.568$\pm$0.063 \\
PCL (ECCV2024)   & OS-Prompt++~\cite{kim2024one}
& 0.743$\pm$0.104 & 0.686$\pm$0.072
& 0.697$\pm$0.032 & 0.655$\pm$0.032
& 0.623$\pm$0.005 & 0.565$\pm$0.023 \\
Arch-CL (CVPR2024)   & MoE-Adapters~\cite{yu2024boosting}
& 0.835$\pm$0.001 & 0.742$\pm$0.001
& 0.747$\pm$0.004 & 0.694$\pm$0.013
& 0.564$\pm$0.002 & 0.478$\pm$0.006 \\
PCL (CVPR2023)   & Coda-Prompt~\cite{smith2023coda}
& 0.682$\pm$0.022 & 0.646$\pm$0.022
& 0.721$\pm$0.001 & 0.697$\pm$0.023
& 0.663$\pm$0.015 & 0.557$\pm$0.061\\
PCL (ECCV2022)   & Dual-prompt~\cite{wang2022dualprompt}
& 0.363$\pm$0.044 & 0.185$\pm$0.002
& 0.435$\pm$0.003 & 0.222$\pm$0.012
& 0.604$\pm$0.042 & 0.259$\pm$0.001 \\
Reh-CL (NIPS2020) & DER++~\cite{buzzega2020dark}
& 0.531$\pm$0.213 & 0.442$\pm$0.253
& 0.609$\pm$0.152 & 0.567$\pm$0.284
& 0.681$\pm$0.043 & 0.612$\pm$0.072 \\
Reg-CL (ICML2018) & Online EWC~\cite{schwarz2018progress}
& 0.746$\pm$0.003 & 0.695$\pm$0.003
& 0.702$\pm$0.002 & 0.708$\pm$0.002
& 0.698$\pm$0.006 & 0.653$\pm$0.004 \\
\midrule
- & \textbf{UniPrompt-CL$^\dagger$}
& \textbf{0.849}$\pm$0.011 & \textbf{0.761}$\pm$0.002
& \textbf{0.772}$\pm$0.032 & \textbf{0.723}$\pm$0.042
& \textbf{0.701}$\pm$0.005 & \textbf{0.656}$\pm$0.001 \\
\bottomrule
\end{tabular}
}
\end{table*}

\begin{table*}[t]
\centering
\small
\caption{Stage-wise comparison of catastrophic forgetting during continual learning.
Red indicates performance on the current training dataset, blue indicates performance on
previously learned datasets, and black indicates performance on unseen future datasets.
The symbol $\dagger$ denotes the proposed model.}
\label{Table_stagewise}
\resizebox{\textwidth}{!}{%
\begin{tabular}{ll cc cc cc   cc cc cc}
\toprule
\multicolumn{2}{c}{}
  & \multicolumn{12}{c}{\textbf{Evaluation}} \\
\cmidrule(l){3-14}
\multicolumn{2}{c}{}
  & \multicolumn{6}{c}{\textbf{OS-Prompt++ (Reference-ViT training)}}
  & \multicolumn{6}{c}{\textbf{UniPrompt-CL$^\dagger$ (Single-pass training)}} \\
\cmidrule(lr){3-8} \cmidrule(l){9-14}
\textbf{Training} & \textbf{Dataset}
  & \multicolumn{2}{c}{\textbf{APTOS}}
  & \multicolumn{2}{c}{\textbf{DDR}}
  & \multicolumn{2}{c}{\textbf{DRD}}
  & \multicolumn{2}{c}{\textbf{APTOS}}
  & \multicolumn{2}{c}{\textbf{DDR}}
  & \multicolumn{2}{c}{\textbf{DRD}} \\
\cmidrule(lr){3-4} \cmidrule(lr){5-6} \cmidrule(lr){7-8}
\cmidrule(lr){9-10} \cmidrule(lr){11-12} \cmidrule(l){13-14}
 &
  & Acc & F1 & Acc & F1 & Acc & F1
  & Acc & F1 & Acc & F1 & Acc & F1 \\
\midrule
Stage 1 & APTOS
  & \textcolor{red}{0.868} & \textcolor{red}{0.753}
  & 0.565 & 0.474
  & 0.409 & 0.354
  & \textcolor{red}{0.901} & \textcolor{red}{0.767}
  & 0.601 & 0.447
  & 0.453 & 0.381 \\
Stage 2 & DDR
  & \textcolor{blue}{0.707} & \textcolor{blue}{0.638}
  & \textcolor{red}{0.797} & \textcolor{red}{0.748}
  & 0.508 & 0.413
  & \textcolor{blue}{0.866} & \textcolor{blue}{0.663}
  & \textcolor{red}{0.878} & \textcolor{red}{0.844}
  & 0.636 & 0.534 \\
Stage 3 & DRD
  & \textcolor{blue}{0.743} & \textcolor{blue}{0.686}
  & \textcolor{blue}{0.697} & \textcolor{blue}{0.655}
  & \textcolor{red}{0.623} & \textcolor{red}{0.565}
  & \textcolor{blue}{0.849} & \textcolor{blue}{0.761}
  & \textcolor{blue}{0.772} & \textcolor{blue}{0.723}
  & \textcolor{red}{0.701} & \textcolor{red}{0.656} \\
\bottomrule
\end{tabular}%
}
\end{table*}

\textbf{FAA / FAF.}
In the loss-term ablation (Fig.~4) we additionally report FAA, the
final average accuracy, and FAF, the final average F1-score, both measured after the last
stage has been trained:
\begin{equation}
\mathrm{FAA} = \frac{1}{T}\sum_{j=1}^{T} R_{T,j},
\end{equation}
where $R_{T,j}$ is the accuracy on stage $j$ after completing all $T$ stages, and FAF is
defined analogously using F1-score as the base performance measure. Unlike AvgACC, which
averages over the whole training trajectory, FAA/FAF summarize only the final-model
performance across stages.

\begin{equation}
\mathrm{CARA}_{p}
= \frac{\mathrm{AvgACC}\times\bigl(1-\mathrm{AvgF}\bigr)}{\mathrm{GFLOPs}^{\,p}},
\qquad p\in[0,1].
\end{equation}

\textbf{Symbols:}
$\mathrm{AvgACC}$ is the average performance, $\mathrm{AvgF}$ is the average forgetting,
$1-\mathrm{AvgF}$ is the retention term, $\mathrm{GFLOPs}$ is the computational cost, and
$p$ controls the strength of the compute penalty.

\textbf{Explanation:}
Cost-Adjusted Retained Accuracy (CARA) combines performance and stability into a retained
performance score and normalizes it by a power of the compute cost, thereby summarizing
the trade-off among accuracy, forgetting, and efficiency in a single value.

\textbf{Meaning:}
A larger CARA value indicates higher performance, lower forgetting, and reasonable
computational cost simultaneously. We report this metric here for completeness
only. Because it normalizes retained accuracy by the very quantity our method
minimizes, we do not use it as evidence of superiority in the main paper, where
accuracy, forgetting, and training cost are reported separately instead. None of
the conclusions in the main paper depends on CARA.

\begin{figure}[t]
\centering
\includegraphics[width=\columnwidth]{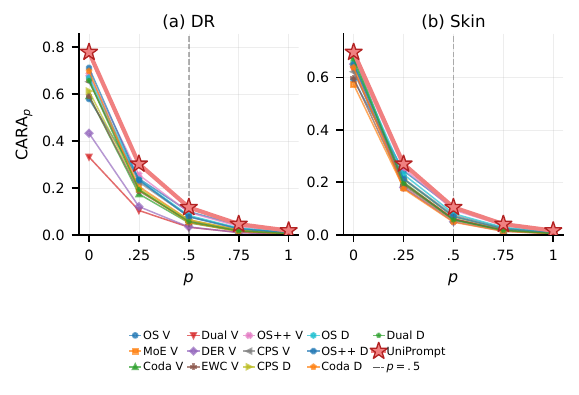}
\caption{Sensitivity of $\mathrm{CARA}_{p}$ over
$p \in \{0,0.25,0.5,0.75,1.0\}$ on the DR and skin cancer
benchmarks. UniPrompt-CL is highlighted with star markers and the dashed
vertical line indicates $p=0.5$. V and D denote ViT-B and DINOv2-B,
respectively.}
\label{fig:cara_sensitivity}
\end{figure}

\begin{table}[t]
\centering
\scriptsize
\setlength{\tabcolsep}{4pt}
\renewcommand{\arraystretch}{0.92}
\caption{Auxiliary $\mathrm{CARA}_{0.5}$ values for every configuration in
Table~1, retained here for completeness. $*$ marks a DINOv2-B
backbone and $\dagger$ the proposed model. AvgACC, AvgF, and GFLOPs, from which
these values are computed, are reported separately in Table~1; the
main paper draws its conclusions from those quantities rather than from this
metric.}
\label{Table_cara}
\begin{tabular}{lccc}
\toprule
\textbf{Method} & \textbf{GFLOPs}$\downarrow$ &
\multicolumn{2}{c}{$\mathbf{CARA}_{0.5}\uparrow$} \\
\cmidrule(lr){3-4}
 & & DR & Skin Cancer \\
\midrule
OS-Prompt~\cite{kim2024one}            & 34.26  & 0.099 & 0.101 \\
MoE-Adapters~\cite{yu2024boosting}     & 105.64 & 0.064 & 0.056 \\
Coda-Prompt~\cite{smith2023coda}       & 134.33 & 0.051 & 0.059 \\
DualPrompt~\cite{wang2022dualprompt}   & 105.05 & 0.032 & 0.061 \\
DER++~\cite{buzzega2020dark}           & 168.02 & 0.033 & 0.050 \\
Online EWC~\cite{schwarz2018progress}  & 100.62 & 0.059 & 0.060 \\
OS-Prompt++~\cite{kim2024one}          & 51.12  & 0.095 & 0.095 \\
CPS-Prompt~\cite{lim2026critical}      & 70.29  & 0.078 & 0.074 \\
\midrule
CPS-Prompt$^*$                         & 92.69  & 0.064 & 0.068 \\
OS-Prompt$^*$                          & 66.42  & 0.082 & 0.081 \\
OS-Prompt++$^*$                        & 86.68  & 0.076 & 0.070 \\
Coda-Prompt$^*$                        & 175.12 & 0.053 & 0.048 \\
DualPrompt$^*$                         & 137.93 & 0.056 & 0.057 \\
\midrule
\textbf{UniPrompt-CL}$^{\dagger*}$     & 44.17  & \textbf{0.117} & \textbf{0.105} \\
\bottomrule
\end{tabular}
\end{table}

\textbf{Sensitivity to the compute penalty.}
Sup.~Table~\ref{Table_cara} lists $\mathrm{CARA}_{0.5}$ for every configuration
of Table~1. The exponent $p$ controls how strongly computation is
penalized: at $p=0$ the
metric reduces to retained accuracy with no penalty, whereas at $p=1$ it applies
a full inverse-GFLOPs penalty. Sup.~Fig.~\ref{fig:cara_sensitivity}
evaluates $p \in \{0,0.25,0.5,0.75,1.0\}$ on both benchmarks. On DR UniPrompt-CL ranks first at every value of $p$, including $p=0$,
where the compute term is absent by construction; on skin cancer it ranks first
up to $p=0.5$ and remains competitive under larger penalties. The ordering is
therefore not an artifact of normalizing by GFLOPs. We record this here rather
than in the main paper because, as noted above, the main conclusions are drawn
from accuracy, forgetting, and training cost reported separately.

\section{Per-Dataset Final-Stage Results}
\label{sec:appendix_perdataset}
Table~\ref{Table1} reports the final-stage accuracy and F1-score of each method on the DR datasets (APTOS, DDR, and DRD) after completing the full
APTOS~$\rightarrow$~DDR~$\rightarrow$~DRD continual learning sequence, complementing the aggregated metrics in Table~1 of the main paper. Each baseline is evaluated in its original backbone configuration, while the backbone-controlled comparison (ViT-B vs. DINOv2-B) is provided in Table~1.

The per-dataset results provide a detailed view of the final retained performance
after continual learning. Because the baselines here use their original backbones
while UniPrompt-CL uses DINOv2-B, this table is not backbone-controlled and
should be read alongside Table~1, which is. Within that caveat,
UniPrompt-CL obtains the highest accuracy and F1-score on all three datasets,
indicating that the proposed unified prompt pool and few-prompt expansion strategy improve not only the aggregated continual learning metrics but also the final
performance retained on each individual domain. In particular, UniPrompt-CL shows clear gains on the earlier APTOS and DDR stages, where forgetting is typically more pronounced after learning subsequent domains. On the final DRD stage, the performance margin over Online EWC is smaller, but UniPrompt-CL still obtains the best Acc and F1, suggesting that the proposed method preserves previously learned knowledge while maintaining competitive
plasticity for the newly learned domain.

\section{Additional Stage-Wise Forgetting Analysis}
\label{sec:appendix_stagewise}

Table~\ref{Table_stagewise} provides a stage-wise breakdown of performance across the
APTOS~$\rightarrow$~DDR~$\rightarrow$~DRD sequence for UniPrompt-CL and OS-Prompt++, the
strongest prompt-based baseline. Diagonal entries (\textcolor{red}{red}) report
current-stage performance, lower-triangle entries (\textcolor{blue}{blue}) report retained
performance on previously learned datasets, and upper-triangle entries (black) report
zero-shot performance on as-yet-unseen future datasets.

The stage-wise results show that UniPrompt-CL consistently retains higher performance on
previously learned datasets while maintaining strong performance on the current stage. After
learning DDR at Stage 2, UniPrompt-CL retains substantially higher APTOS accuracy than
OS-Prompt++ (0.866 vs. 0.707), indicating reduced forgetting immediately after the first
domain shift. After completing the full sequence at Stage 3, UniPrompt-CL also preserves
higher performance on both previous domains, improving APTOS Acc/F1 from 0.743/0.686 to
0.849/0.761 and DDR Acc/F1 from 0.697/0.655 to 0.772/0.723 compared with OS-Prompt++.
These retained-performance gains are achieved while UniPrompt-CL preserves a single-backbone
single-pass design, whereas OS-Prompt++ requires an additional reference-ViT pass during
training. The results therefore support that the unified prompt pool and few-prompt
expansion strategy improve resistance to catastrophic forgetting without sacrificing
computational efficiency.

\section{GenAI Usage Disclosure}
\label{A_A}

Although the authors carried out the overall code development and manuscript preparation,
the conversion of the written manuscript into LaTeX, its proofreading, and translation
were performed using OpenAI's ChatGPT, and code refactoring was assisted by Cursor.

\clearpage
\end{cbunit}
\end{document}